\documentclass[conference,a4paper]{IEEEtran}
\IEEEoverridecommandlockouts
\usepackage[space]{cite}
\usepackage{amsmath,amssymb,amsfonts}
\usepackage{algorithmic}
\usepackage{graphicx}
\usepackage{textcomp}
\usepackage{xcolor}
\def\BibTeX{{\rm B\kern-.05em{\sc i\kern-.025em b}\kern-.08em
    T\kern-.1667em\lower.7ex\hbox{E}\kern-.125emX}}

\usepackage[breaklinks=true,colorlinks,bookmarks=false]{hyperref}

\usepackage[labelformat=simple,subrefformat=simple]{subcaption}

\begin{document}

\title{Online pre-training with long-form videos\\
}

\author{
\IEEEauthorblockN{Itsuki Kato, Kodai Kamiya, Toru Tamaki}
\IEEEauthorblockA{\textit{Nagoya Institute of Technology}, Japan \\
toru.tamaki@nitech.ac.jp}
}

\maketitle

\begin{abstract}
In this study, we investigate the impact of online pre-training with continuous video clips. We will examine three methods for pre-training (masked image modeling, contrastive learning, and knowledge distillation), and assess the performance on downstream action recognition tasks. As a result, online pre-training with contrast learning showed the highest performance in downstream tasks. Our findings suggest that learning from long-form videos can be helpful for action recognition with short videos.

\end{abstract}


\section{Introduction}

Deep learning models for video understanding typically rely on locally stored files, which implies offline processing. However, research is also advancing in the field of online video recognition. In typical offline training, frames and videos are randomly chosen for model update. Yet, for long-form videos such as streams, random selection and precomputation of features are not practical. As a result, previous research has focused on creating efficient models and learning downstream tasks through causal inference, which does not rely on future frames.
However, when causally learning a model from a long video, using continuous clips of video frames may result in redundancy, potentially degrading performance.
In this study, we investigate the impact of pretraining with continuous video clips. We will examine three methods: Masked Image Modeling (MIM), contrastive learning, and knowledge distillation.

\section{Online pre-training}

\subsection{Video clip}

Consider a video dataset denoted as
$V=\{v_{o},v_{1},...,v_{N}\}$,
each $v_{i} \in \mathbb{R}^{T_{i}\times{3}\times{H}\times{W}}$ represents a video with $T_{i}$ frames of
height $H$ and width $W$.
From each video $v_{i} \in \mathbb{R}^{T_{i}\times{3}\times{H}\times{W}}$, we define a continuous video clip of $T$ frames with a stride $s$;
\begin{equation}
x_{i,n} = v_{i}[nsT : (n + 1)sT : s, :, :, :]
\in \mathbb{R}^{T \times 3 \times H \times W},
\end{equation}
with a video clip index $n$
that ranges from 0 to $n_i = \lfloor \frac{T_{i}}{sT}\rfloor - 1$. 

In this study, we used video clips, sequentially extracted from lengthy videos, to pre-train the model. These clips are represented as $x_{0, 0}, x_{0, 1}, \ldots, x_{0, n_0}, x_{1, 0}, \ldots, x_{1, n_1}, \ldots x_{N, n_N}$. To simplify the notation, we denote them by $x$ when the indices $i, n$ are not significant.

\subsection{Self-supervised learning}

In this study, we discuss three methods for online pre-training. The first is Masked Image Modeling (MIM) \cite{Zhang_arXiv2022_survey_MAE,Tong_NEURIPS2022_VideoMAE}. Here, the patches for Vision Transformer (ViT) \cite{Dosovitskiy_ICLR2021_ViT_Vision_transformer} are randomly masked, and their contents are predicted. 
The second is contrastive learning that closely aligns the embeddings by two encoders from a single training sample.
The third is knowledge distillation \cite{Hu_arXiv2023_survey_Teacher_Student} with teacher-student learning. This involves training the student model to match the output of the pre-trained teacher model.

\section{Experimental results}

\subsection{Dataset}

\subsubsection{Pre-training}
AVA-Actions\cite{Gu_2018CVPR_AVA-Actions} is a dataset that includes 235 training videos and 64 evaluation videos. Each video spans 15 minutes. In this study, we do not use annotated labels.

\subsubsection{Downstream Tasks}
HMDB51\cite{Kuehne_ICCV2011_HMDB51} is an action recognition dataset. It consists of 51 categories and 6766 videos, with 3570 videos for training and 1530 videos for evaluation.
UCF101\cite{Soomro_arXiv2012_UCF101} is another action recognition dataset. It comprises 101 categories and 13320 videos, with 9537 videos for training and 3783 videos for evaluation.

\subsection{Video clip preparation for pretraining}

In our experiment, we compare two approaches of clip preparation;

\begin{itemize}
\item Random clip sampling: we divided each videos of AVA-Actions into 10-second trimmed video segments. For training, we randomly select one of these trimmed videos and retrieve $T$ frames with a stride of $S$. We then shuffle the retrieved $T$ frames to create a video clip.
\end{itemize}

\begin{itemize}
\item Sequential clip sampling: we extracted frames in sequence from the beginning of videos. From the initial frame of a randomly selected video of AVA-Actions, we extract $T$ consecutive frames with a stride of $S$ to create a video clip.
\end{itemize}

In both methods, we normalize the pixel values of each frame in the created clips to have a mean of 0 and a standard deviation of 1, and resize the pixels to $224\times224$ following a common procedure.

\subsection{Settings}

\subsubsection{Pre-training}
During pre-training, we compare random clips with sequential clips. 
Both types of clips contain $T=16$ frames. 
The random clip has a stride of $S=8$, whereas the stride of the sequential clip varies at $S=16,32,64$ in our experiments.

For MIM, we used VideoMAE \cite{Tong_NEURIPS2022_VideoMAE} and
set the mask rate established at 50\%.
For contrastive learning, we used MoCo \cite{He_CVPR2020_MoCo,Pan_CVPR2021_VideoMoCo} with X3D-M \cite{Feichtenhofer_2020CVPR_X3D}, pre-trained on Kinetics400 \cite{kay_arXiv2017_kinetics400}, as encoder. 
For teacher-student learning, both models are X3D-M, but the teacher model is pre-trained on Kinetics-400 \cite{kay_arXiv2017_kinetics400}, while the student model is scratch.

\subsubsection{Downstream task}
we assess the pre-trained models using the three aforementioned pre-train methods based on their performance in action recognition tasks.
We used a standard configuration with training of 30 epochs for
HMDB51\cite{Kuehne_ICCV2011_HMDB51}
and 10 epochs for UCF101\cite{Soomro_arXiv2012_UCF101}.

\begin{figure}[t]
  \centering

    \subcaptionbox{\label{Fig:MIM_rec_val}}{%
        \includegraphics[width=0.48\linewidth]{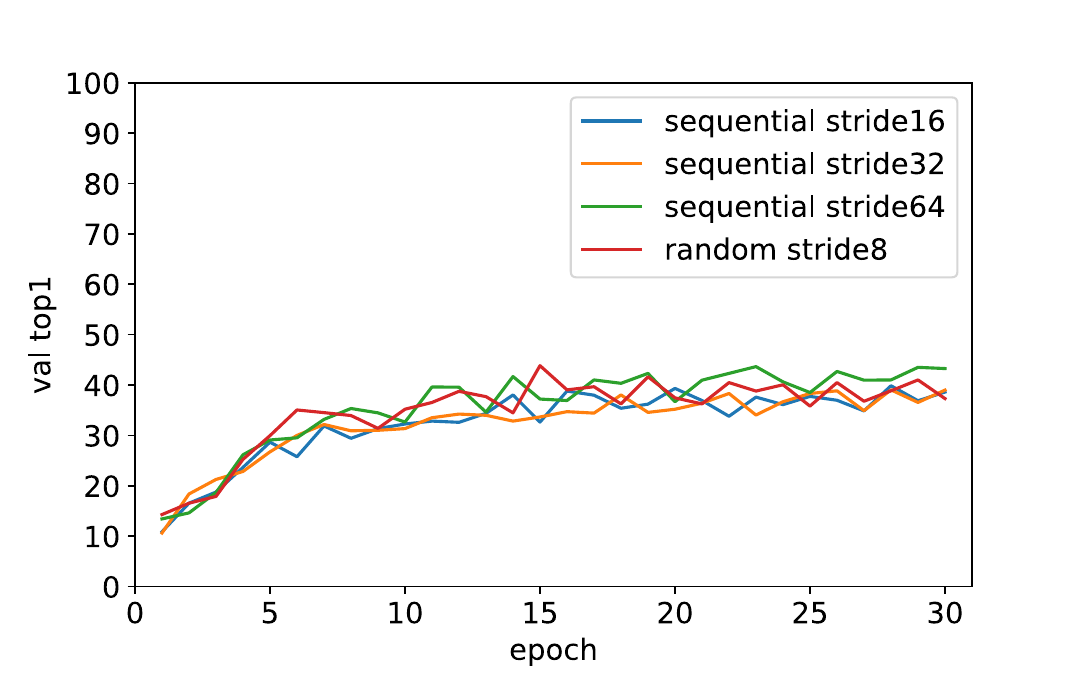}
    }
    \hfill
    \subcaptionbox{\label{Fig:MIM_rec_val_UCF}}{%
        \includegraphics[width=0.48\linewidth]{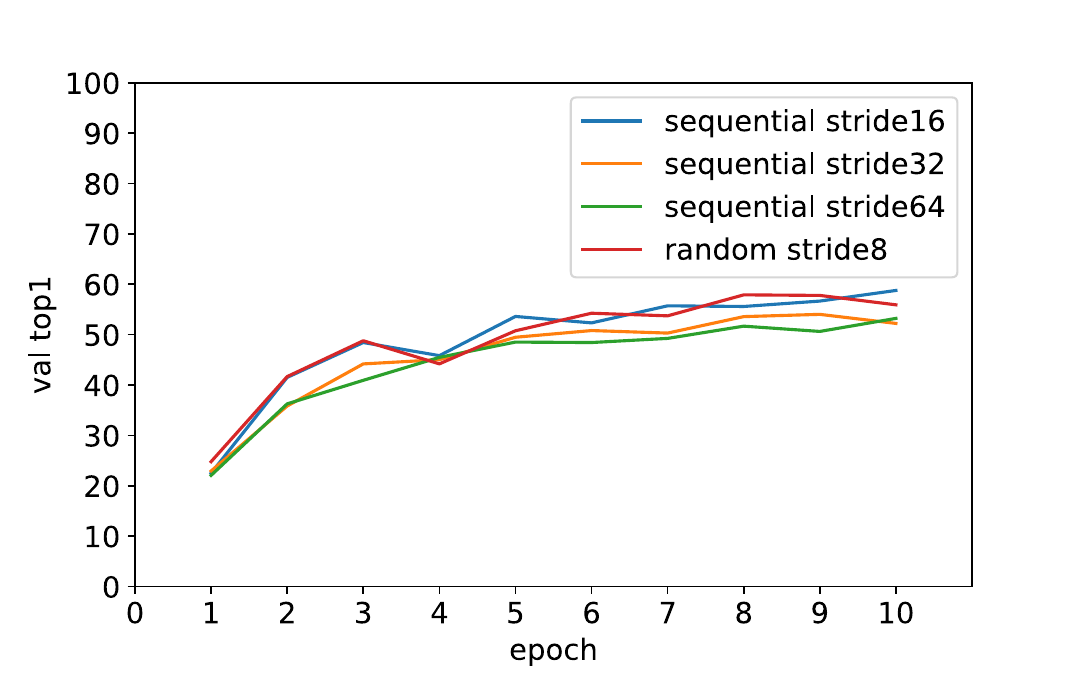}
    }

    \medskip

    \subcaptionbox{\label{Fig:moco_rec_val}}{%
        \includegraphics[width=0.48\linewidth]{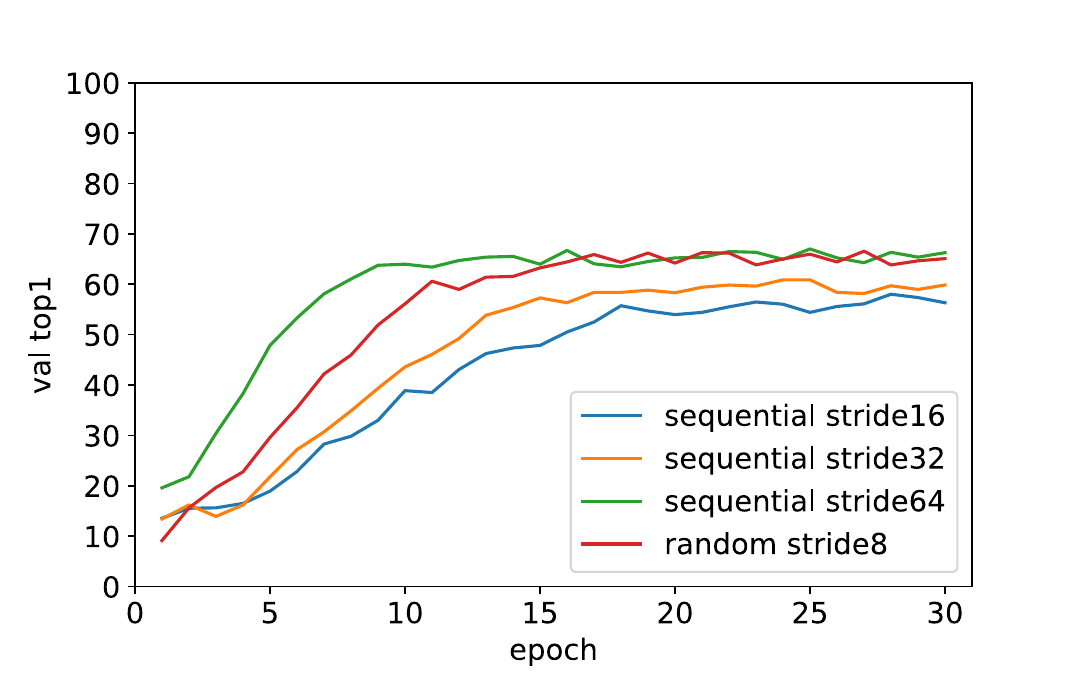}
    }
    \hfill
    \subcaptionbox{\label{Fig:moco_rec_val_UCF}}{%
        \includegraphics[width=0.48\linewidth]{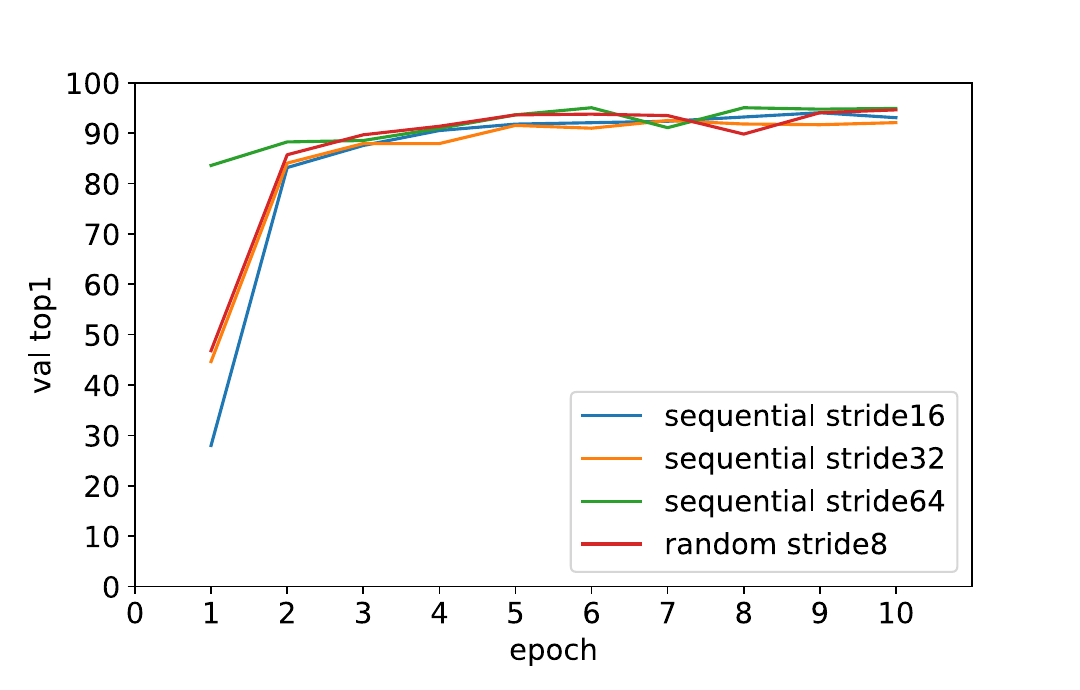}
    }

    \medskip

    \subcaptionbox{\label{Fig:TSL_rec_val}}{%
        \includegraphics[width=0.48\linewidth]{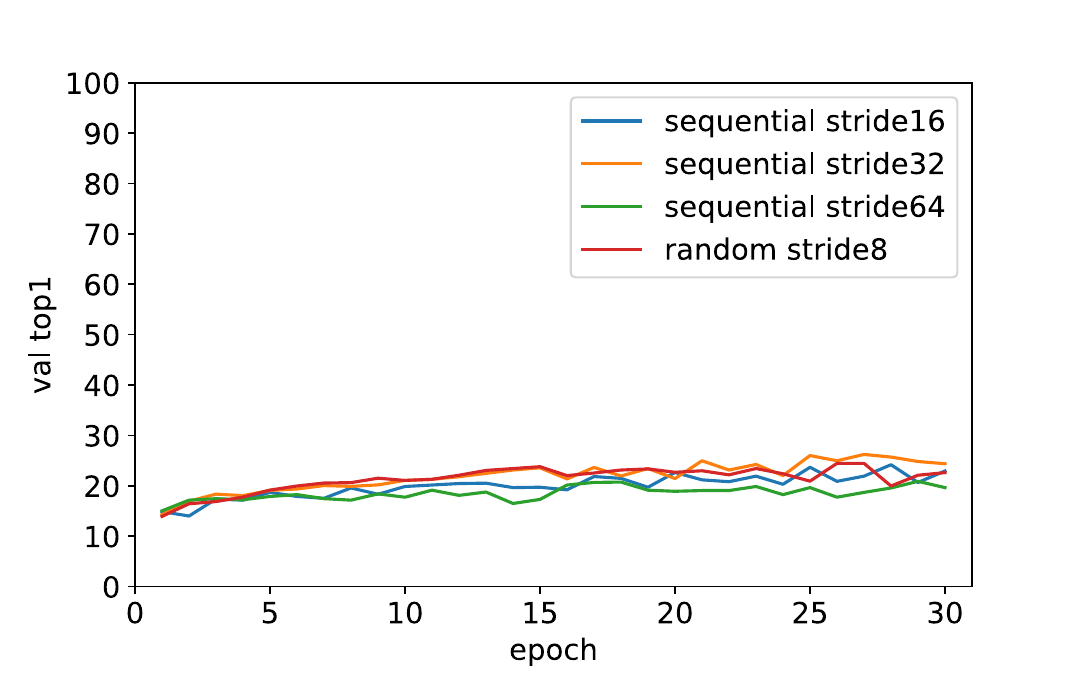}
    }
    \hfill
    \subcaptionbox{\label{Fig:TSL_rec_val_UCF}}{%
        \includegraphics[width=0.48\linewidth]{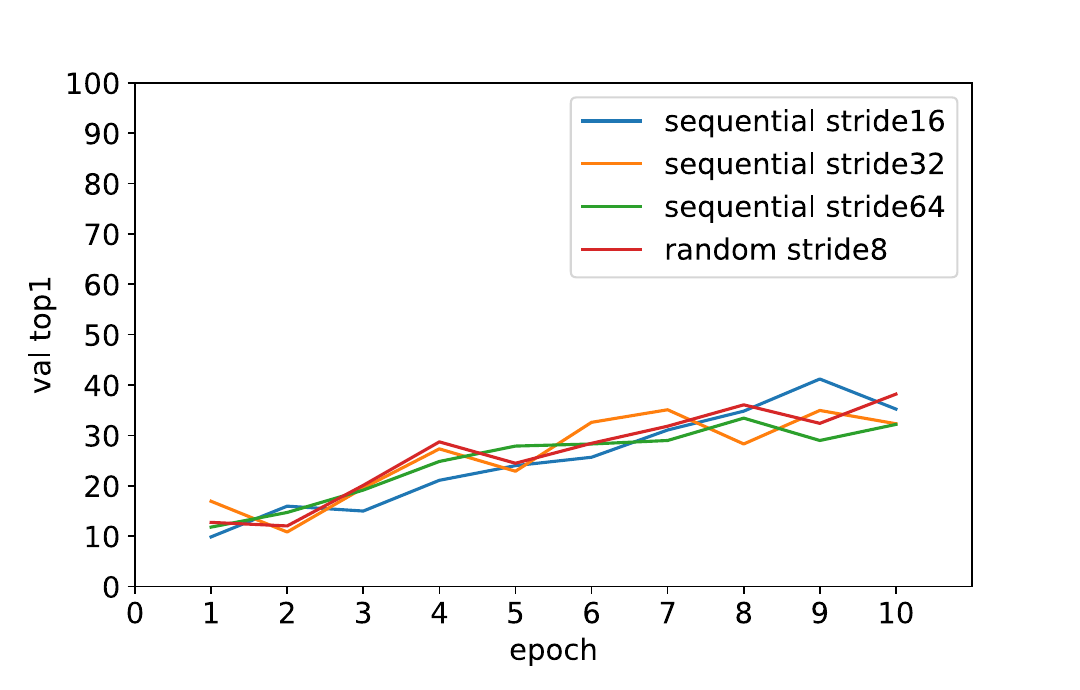}
    }

  \caption{Top-1 performance of the downstream tasks with
  \subref{Fig:MIM_rec_val}\subref{Fig:MIM_rec_val_UCF} MIM,
  \subref{Fig:moco_rec_val}\subref{Fig:moco_rec_val_UCF} contrastive learning, and
  \subref{Fig:TSL_rec_val}\subref{Fig:TSL_rec_val_UCF} techer-student learnig.
  left: HMDB51, right: UCF101.
  }
  \label{fig:result}

\end{figure}

\subsection{Result}

\subsubsection{MIM}

In Figures
\ref{fig:result}\subref{Fig:MIM_rec_val}\subref{Fig:MIM_rec_val_UCF}, the
performances of sequential clips are comparable to those of random clips;
however, the similar performance of all settings in the downstream tasks suggests that introducing stride has helped mitigate overfitting of sequential clips.

\subsubsection{Contrastive learning}

As shown in Figures
\ref{fig:result}\subref{Fig:moco_rec_val}\subref{Fig:moco_rec_val_UCF},
there is a positive correlation between stride and performance for sequential clips. This indicates that a larger stride in pretraining reduces the risk of overfitting.
Of the three compared methods, contrastive learning showed the best performance in downstream tasks. 
Although the pre-training dataset is commonly used for action detection \cite{Gritsenko_arXiv2023_STAR}, our findings suggest that learning from long-form videos can be helpful for action recognition tasks with short videos.

\subsubsection{Teacher-student learning}

As in Figures
\ref{fig:result}\subref{Fig:TSL_rec_val}\subref{Fig:TSL_rec_val_UCF},
the performances are significantly lower than expected for these datasets, indicating that training has not been successful.
This is likely due to insufficient distillation from the teacher model during pre-training.

\section{Conclusion}

In this paper, we explore the online pretraining methods for long-form videos with sequential clip sampling. As a result, online pretraining with contrast learning showed the highest performance in downstream tasks. In future studies, exploring techniques to minimize the correlation between consecutive clips could be beneficial for sequential clip sampling.

\section*{Acknowledgements}
This work was supported in part by JSPS KAKENHI Grant Number JP22K12090.

\bibliography{mybib,all}
\bibliographystyle{ieeetr}

\end{document}